\documentclass[11pt]{article}

\usepackage[final]{acl}

\usepackage{times}
\usepackage{latexsym}
\usepackage{hyperref}
\usepackage{booktabs}
\usepackage{subcaption}
\usepackage{adjustbox}
\usepackage{listings}
\usepackage{xcolor}

\usepackage[T1]{fontenc}
\usepackage[utf8]{inputenc}
\usepackage{comment}
\usepackage{microtype}

\usepackage{inconsolata}

\usepackage{graphicx}

\title{Grounded Satirical Generation with RAG}
\author{
Oona Itkonen\thanks{Equal contribution} \and 
Yuxin Su\footnotemark[1] \and 
Linyao Du\footnotemark[1] \and 
Ona De Gibert \\
University of Helsinki\\
}

\begin{document}
\maketitle
\begin{abstract}
Humor generation remains challenging task for Large Language Models (LLMs), due to their subjective nature. We focus on satire, a form of humor strongly shaped by context. In this work, we present a novel pipeline for grounded satire generation that uses Retrieval-Augmented Generation (RAG) over current news to produce satirical dictionary definitions in the Finnish context. We also introduce a new task-specific evaluation framework and annotate 100 generated definitions with six human annotators, enabling analysis across multiple experimental conditions, including cultural background, source-word type, and the presence or absence of RAG.
Our results show that the generated definitions are perceived as more political than humorous. Both topic-based word selection and RAG improve the political relevance of the outputs, but neither yields clear gains in humor generation. In addition, our LLM-as-a-judge evaluation of five state-of-the-art models indicates that LLMs correlate well with human judgments on political relevance, but perform poorly on humor. We release our code and annotated dataset to support further research on grounded satire generation and evaluation.
\end{abstract}

\section{Introduction}

Humor is a fundamental aspect of human nature, yet defining it remains a persistent challenge \cite{bardon2005philosophy,larkin2017overview}. Given this conceptual complexity, it is unsurprising that Large Language Models (LLMs) also struggle to reliably interpret and generate humor.
LLMs can detect humor with relatively well, but generating it is still an unsolved problem. 

In this work, we focus on satire generation and evaluation, an even more challenging task, as satire constitutes a nuanced cultural-specific form of humor, the interpretation of which depends heavily on a shared social, political, and historical context \cite{stinson2019satire}.
We adopt the definition of \textit{Satire} from the Cambridge Dictionary\footnote{\url{https://dictionary.cambridge.org/dictionary/english/satire}}: \textit{a way of criticizing people or ideas in a \textbf{humorous} way, especially in order to make a \textbf{political point}}. Accordingly, two core components of satire are humor and political significance.

\begin{figure*}[ht]
    \centering
    \includegraphics[width=\linewidth]{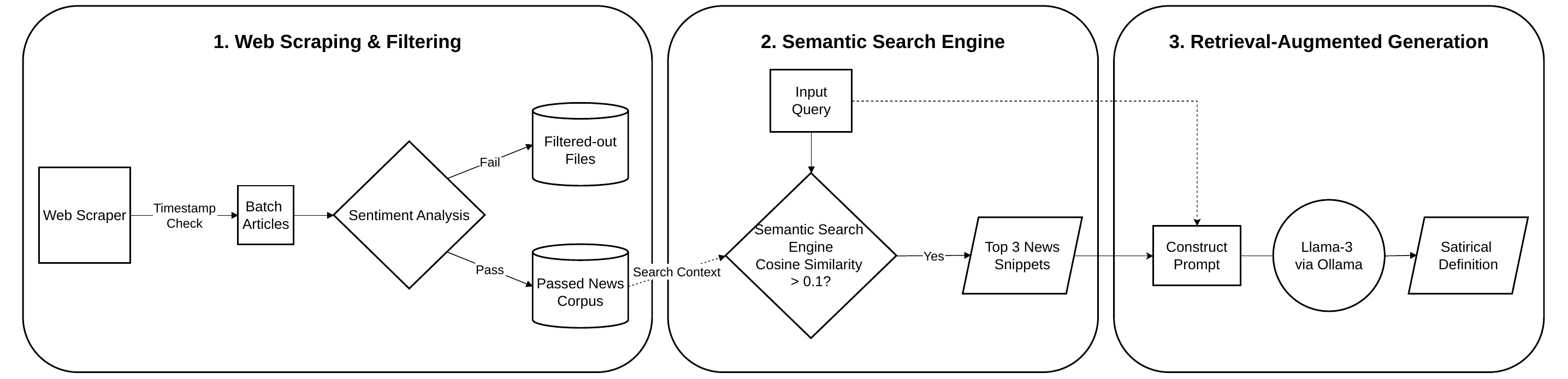}
    \caption{Overview of our generation pipeline for satirical dictionary definitions with RAG.}
    \label{fig:pipeline}
\end{figure*}

For generation, we present a novel method based on Retrieval-Augmented Generation (RAG) \citep{lewis2021retrievalaugmentedgenerationknowledgeintensivenlp} to produce satire from news content in the form of satirical dictionary definitions. Because satire is highly dependent on cultural and regional context, we restrict our study to the Finnish setting. As source data, we use English-language news articles published on the website of the Finnish public broadcaster Yle\footnote{\url{https://yle.fi/}}. Our pipeline consists of a scraper with sentiment analysis and timestamp-based filtering, a semantic search engine, a topic modeling component for selecting candidate words, and a final RAG-based generation stage.

For evaluation, we develop our own framework, motivated by the fact that humor evaluation is inherently difficult and lacks standardized evaluation practices \cite{hamalainen-alnajjar-2021-human}.  We present evaluation results of based on two methods: human annotation and LLM-based judgment. We evaluate five mid-size state-of-the-art LLMs.
We aim to answering the following Research Questions (RQ):
\begin{itemize}
\item RQ1: To what extent are the generated definitions humorous and politically meaningful?
\item RQ2: To what extent is successful satire generation dependent on cultural context?
\item RQ3: How does the choice of the candidate word affect the quality of the generated definitions?
\item RQ4: Does RAG improve the quality of generated satirical definitions?
\item RQ5: Can LLMs serve as reliable evaluators of satirical content?
\end{itemize}

In addition, we present a webpage application to showcase and run our generation pipeline. Furthermore, we release our annotated corpus to the research community and make our scripts publicly available for reproducibility\footnote{\url{https://github.com/dlylinyao/ONLY/tree/CHUM}}.

\section{Related Work}
One of the first tasks related to humor in NLP is \textbf{humor detection}.
Early work in computational humor focused on template-based methods \cite{chandrasekaran2016we} and classical supervised machine learning approaches relying on static features \cite{de2015humor}. More recently, transformer-based and LLM-based methods have been proposed. Recent research has examined the generalizability of humor detection models \cite{baranov-etal-2023-told}, as well as related tasks such as irony detection \cite{ortega2023cross,tomas2023transformer,lin2024augmenting} and \textbf{humor understanding} \cite{hessel-etal-2023-androids,hwang2025bottlehumor}, in which an LLM is prompted to explain why a given text is humorous.

\textbf{Humor generation} has also attracted attention for many years \cite{stock2005hahacronym}. More recently, \citet{jentzsch-kersting-2023-chatgpt} investigated the use of ChatGPT for joke generation and found that the model tends to produce repetitive outputs. Similarly, \citet{sakabe2025assessing} used LLMs to generate \textit{Oogiri}, a form of Japanese improvisational comedy. However, the field continues to observe that, while LLMs perform relatively well on humor detection, they remain less effective at humor generation \cite{sakabe2025assessing}.

Some work has also tried to develop improved \textbf{humor evaluation} methods. For example, \citet{romanowski2025punchlines} propose a new metric for evaluating stand-up comedy based on statistical measures. However, humor evaluation remains difficult because no standard evaluation practices have been established. A recent trend is to design task-specific annotation guidelines, obtain judgments from both human annotators and LLMs, and then analyze the agreement between them \cite{bago2025few,sakabe2025assessing,rivera2026not}. This is the approach we adopt in this work.

Finally, turning specifically to \textbf{satire}, research on satire and LLMs has expanded rapidly in recent years. One line of work focuses on satire detection. For example, \citet{ozturk-etal-2025-make} introduce a dataset of Turkish satirical news and investigate methods for reducing stylistic bias in satire detection.
Another direction has been proposed by \citet{west2019reverse,horvitz-etal-2024-getting}, who study satire by reverse-engineering satirical news headlines and evaluating LLMs’ ability to make them non-satirical or “unfunny.” \citet{dobre2025evaluating} compare AI-generated satire with human-written satirical articles and evaluate the outputs using an LLM-as-a-judge framework. To the best of our
knowledge, we are the first to study satirical generation grounded on latest news using RAG.

\begin{table*}[ht!]
\centering

\begin{minipage}[t]{0.48\textwidth}
  \centering
  \begin{tabular}{ll}
    \toprule
    \multicolumn{2}{l}{\textbf{Q1: Is it funny?}} \\
    \midrule
    \textbf{Score} & \textbf{Explanation} \\
    1 & It is not funny, awkward at most. \\
    2 & It is slightly funny. \\
    3 & It is so funny I laughed. \\
    4 & It is so funny I will tell it to \\
      & someone else. \\
    5 & It is so funny I will laugh if \\
      & I tell it to someone else. \\
    \bottomrule
  \end{tabular}
  \caption{Annotation guideline for Q1: \textit{Is it funny?}}
  \label{tab:Q1}
\end{minipage}
\hfill
\begin{minipage}[t]{0.48\textwidth}
  \centering
  \begin{tabular}{ll}
    \toprule
    \multicolumn{2}{l}{\textbf{Q2: Is it political?}} \\
    \midrule
    \textbf{Score} & \textbf{Explanation} \\
    1 & It is not political at all. \\
    2 & It has some political quality. \\
    3 & It is political on a general level. \\
    4 & It is political and current. \\
    5 & It is political, current and rele- \\
      & vant in Finnish political \\
      & culture. \\
    \bottomrule
  \end{tabular}
  \caption{Annotation guidelines for Q2: \textit{Is it political?}}
  \label{tab:Q2}
\end{minipage}

\end{table*}

\section{Generation}

In this section we present our methodology for grounded generation of satirical dictionary definitions. Our pipeline is based on a web scraper that includes timestamp filtering and sentiment analysis. We select relevant candidate words using topic modeling, find the relevant articles from our data with semantic search and, finally, we use RAG to generate definitions of the words based on the latest news. In Section~\ref{UI}, we present how this pipeline can be run as a web application. Figure \ref{fig:pipeline} presents an overview of our pipeline.

\subsection{Web Scraping}
\label{sec:scraper}
%Linyao, Oona
Given the definition of satire from Cambride Dictonary\footnote{\url{https://dictionary.cambridge.org/dictionary/english}}, satire is often political. 
Therefore, our data should ideally have political content. As news are often about politics, or at least have content that can be interpreted in a political frame, we choose to  use news as our source data. Politics is, however, not a very static field of domain and it evolves quickly to new topics. Therefore, rather than opting for a static news dataset, we built a scraper with BeautifulSoup to extract news articles published in English from the website of the Finnish broadcasting company Yle. Our scraper retrieves articles from all the different categories listed in the Yle website in English\footnote{\url{https://yle.fi/news}}, parsing their metadata. As satire is, in addition to being political on a general level, also depended on culture and region, we restrict our data to this one source, as it is the only open source site that publishes news in English in Finland.

\subsection{Filtering}
\label{sec:filtering}
\subsubsection{Timestamp Filtering}

For the same reason, discussed in section \ref{sec:scraper}, we choose to scrape our data rather than use an existing dataset we want to filter our data based on the timestamps of the articles.  
We chose to use 30 days as the threshold for timestamps, and articles older than a month do not get processed any further.

\subsubsection{Sentiment Analysis}

The Cambridge Dictionary\footnote{\url{https://dictionary.cambridge.org/dictionary/english}} describes satire as a way of criticizing something in a humorous way. However, there are certain restrictions to what kind of topics are commonly considered acceptable for satire. On the other hand, when satire is made based on sensitive topics, the tiniest details in the choice of words can define whether it is interpreted as offensive and inappropriate instead of funny. 

This kind of contextual understanding is something that humans are able to consider. As we are generating satire with an LLM, however, we need to be aware of restrictions that the model has in understanding what is appropriate and what is not. Therefore, to ensure our pipeline doesn't produce any offensive or disturbing content, we implement sentiment analysis to filter out too negative news. Here, we are, assuming that sentiment is a proxy for how "bad" a piece of news is or how sensitive or severe the topics it covers might be.

For the sentiment analysis task, we use the \href{https://huggingface.co/nlptown/bert-base-multilingual-uncased-sentiment}{NLP-Town/bert-base-multilingual-uncased-sentiment} model from Hugging Face. 
We feed the body text of the news articles to the model, and it outputs from one up to five stars, ranging from more negative to more positive.

Our motivation for sentiment analysis is not to filter out everything negative, but rather to ensure the ethics of our output. Therefore, we set the threshold of our sentiment analysis to one, so news that get a label lower than one are discarded.
To obtain a more reliable sentiment score for each news article, we split the articles into batches that fit within the model’s token limit, perform sentiment analysis on each batch, and then compute the mean label scores to obtain the final article-level score.

\subsection{Word Candidate Selection with Topic Modeling}
%Yuxin, Oona
\label{sec:topic}

To generate satire that makes sense in a specific cultural context, we find topics currently discussed in the news with unsupervised topic modeling to automatically extract candidate words from our web-scraped data.

First, we convert the news articles into text embeddings using the \href{https://huggingface.co/sentence-transformers/paraphrase-MiniLM-L12-v2}{paraphrase-multilingual-MiniLM-L12-v2}  
model. We then apply UMAP \cite{mcinnes2020umapuniformmanifoldapproximation} for dimensionality reduction and use BERTopic \cite{grootendorst2022bertopicneuraltopicmodeling} to cluster the articles into distinct news topics. After excluding outlier documents, we extract the most salient keywords from each valid cluster. This words will be used to generate satirical dictionary definitions.

\subsection{Retrieval}

To retrieve relevant news for each input word, we employ a semantic search approach to prioritize contextual meaning over exact keyword matching.
We embed the news articles using the \href{https://huggingface.co/sentence-transformers/all-MiniLM-L6-v2}{all-MiniLM-L6-v2} model. 

For each input, we calculate cosine similarity to find the most relevant news contexts. The system retrieves up to 3 news snippets per input word or phrase, filtering out any results with a similarity score below 0.1. If the search engine finds exact matches of the input, the snippets consist of the immediate contexts around those matches. If the word does not occur in any of the articles, the search engine returns the snippet from the beginning of the article. The snippets are of the size 160 character each, to preserve context most relevant for the input and they always include the timestamp, the category and the title of the article.

\subsection{RAG}
We built a custom RAG system to generate definitions for words based on the news data.
The retrieved snippets are passed to \href{https://huggingface.co/meta-llama/Meta-Llama-3-8B-Instruct}{meta-llama/Meta-Llama-3-8B-Instruct} \cite{grattafiori2024llama} using the ollama library\footnote{\url{https://ollama.com/library/llama3:latest}}, prompted to act as the editor of a “Satirical Dictionary.” To ensure the humor is grounded, the prompt strictly instructs the model to base its definitions solely on the provided news context rather than generic stereotypes. We enforce a cynical tone to highlight the absurdity of the specific news events and limited outputs to 50 words. The exact prompt can be found in Appendix \ref{sec:prompts}.

\begin{figure*}[t]
    \centering
    \begin{subfigure}[t]{0.48\textwidth}
        \centering
        \includegraphics[width=\linewidth]{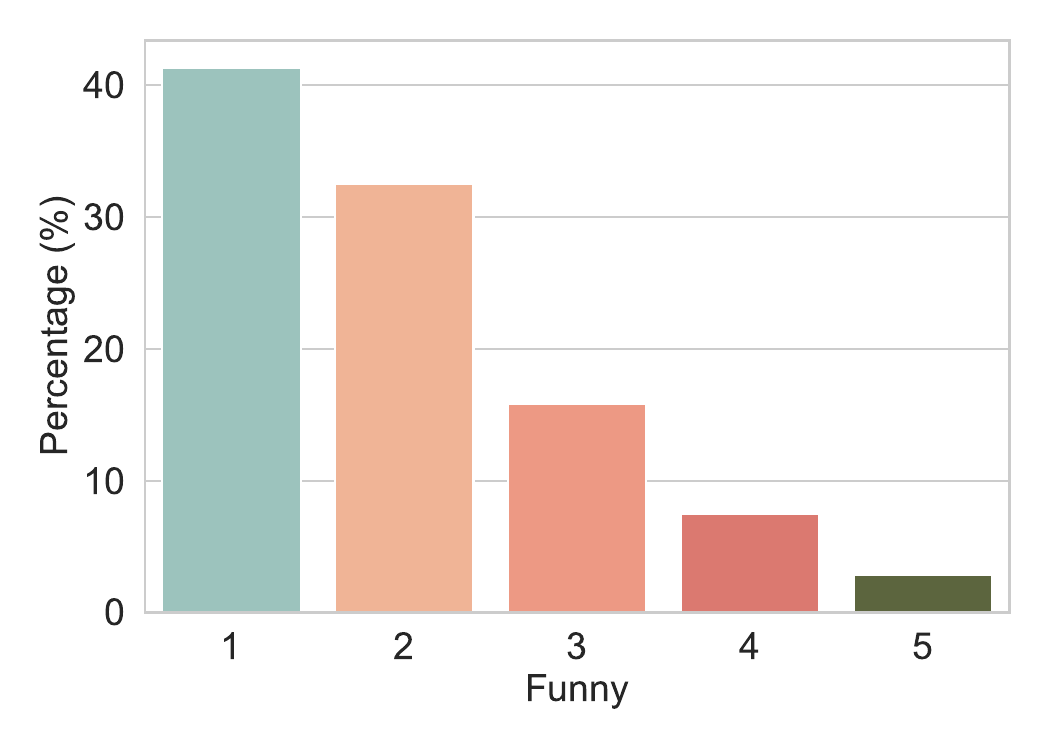}
        \caption{Funny score distribution}
        \label{fig:funny_dist}
    \end{subfigure}\hfill
    \begin{subfigure}[t]{0.48\textwidth}
        \centering
        \includegraphics[width=\linewidth]{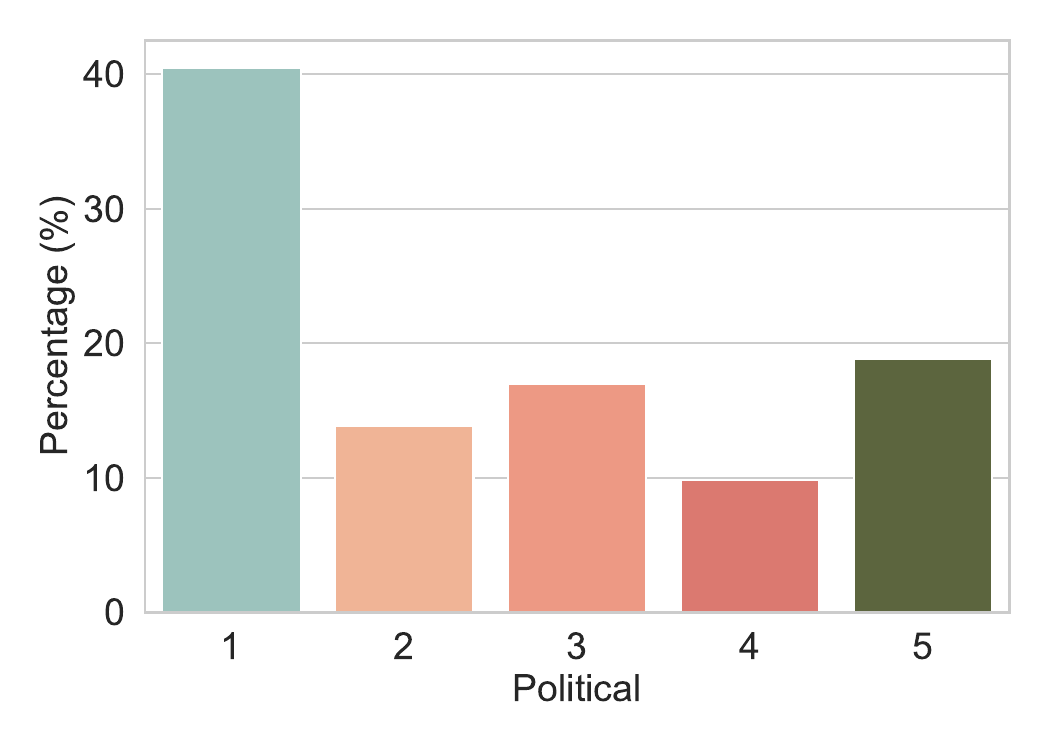}
        \caption{Political score distribution}
        \label{fig:political_dist}
    \end{subfigure}
    \caption{Distribution of absolute quality scores across annotators.}
    \label{fig:score_distributions}
\end{figure*}

\section{Evaluation} 
As the amount of existing research on computational satire is rather limited, and nonexistent when it comes to grounded generation of satire, there is no standard evaluation method that we could apply in this study. We choose to use human evaluation as our evaluation method, and in addition to that, we investigate whether LLMs agree with our human evaluation results using LLMs as judges. We base the choice and design of our evaluation method on practices from creative natural language generation \cite{hamalainen-alnajjar-2021-human} and existing studies on computational satire evaluation \citep{west2019reverse, horvitz-etal-2024-getting, dobre2025evaluating}. 

For the evaluation, we generate definitions for 50 words. 25 of these are drawn from the topics identified in Section~\ref{sec:topic}. The remaining 25 consist of randomly selected English words.  This setup allows us to examine whether news-related words lead to different generated definitions than unrelated random words. 
In addition, for each word we generate two definitions, one with RAG and one without RAG, in order to assess the impact of retrieval augmentation. 
The exact prompts can be found in Appendix \ref{sec:prompts}.  
For generating the defintions, we use news articles scraped on March 3, 2026.

We develop our own annotation guidelines both for human evaluation and LLM judges. Drawing on the Cambridge Dictionary definition of satire, we formulate two questions based on two dimensions: \textit{Q1: Is it funny?}, and \textit{Q2: Is it political?}. Annotators are then asked to rate each definition on both questions using a 1-to-5 Likert scale. For each question, we provide a verbal description of the scores from 1 to 5, as shown in Tables \ref{tab:Q1} and \ref{tab:Q2}.

\begin{table}[]
    \centering
    \begin{tabular}{lrr}
    \toprule
    Annotator Group & Humor & Politics \\
    \midrule
    All &  0.070 & 0.514 \\
    Finnish & 0.053 &  0.646    \\
    International & 0.183 & 0.490 \\
    \bottomrule
    \end{tabular}
    \caption{Inter-annotator agreement of normalized z-scores measured by Krippendorf's $\alpha$ }
    \label{tab:IAA}
\end{table}

\subsection{Human Evaluation}

For the human evaluation, we randomly shuffle the 100 definitions to ensure a blind annotation setup, such that annotators rate each sample without knowing which model generated it or under which experimental condition it was produced.

We employ six annotators to evaluate the 100 definitions. Half of the annotators are Finnish, while the other half come from different cultural backgrounds. This design allows us to examine whether cultural background influences the interpretation of the generated definitions.

\subsection{LLM-as-a-Judge}
\label{sec:llmasajudge}
Human annotation is costly and time-consuming. Recently, LLMs have increasingly been used as automatic evaluators for a variety of NLP tasks. We aim to investigate whether LLMs can reliably evaluate humor and political relevance in satirical definitions.

To this end, we evaluate several open-weight models and instruct them to score definitions according to the same annotation guidelines used in the human evaluation setting described above. Each model is prompted with the evaluation prompt provided in Appendix \ref{sec:prompts}. The models assign scores for humor and political relevance following the same scale as used by human annotators.

We evaluate the following instruction-tuned models of comparable size:
\href{https://huggingface.co/Qwen/Qwen2.5-7B-Instruct}{Qwen/Qwen2.5-7B-Instruct} \cite{ahmed2025qwen}, 
\href{https://huggingface.co/meta-llama/Llama-3.1-8B-Instruct}{meta-llama/Llama-3.1-8B-Instruct} \cite{grattafiori2024llama},
\href{https://huggingface.co/mistralai/Mistral-7B-Instruct-v0.3}{mistralai/Mistral-7B-Instruct-v0.3} \cite{jiang2023mistral7b},  
\href{https://huggingface.co/CohereLabs/aya-expanse-8b}{CohereLabs/aya-expanse-8b} \cite{dang2024aya}
and \href{https://huggingface.co/utter-project/EuroLLM-9B-Instruct}{utter-project/EuroLLM-9B-Instruct} \cite{martins2025eurollm}.

\section{Results}

In this section, we analyze the annotated data (both by humans and LLMs) to address the research questions outlined in the Introduction.

\begin{figure*}[t]
    \centering
    \begin{subfigure}[t]{0.33\textwidth}
        \centering
        \includegraphics[width=\linewidth]
        {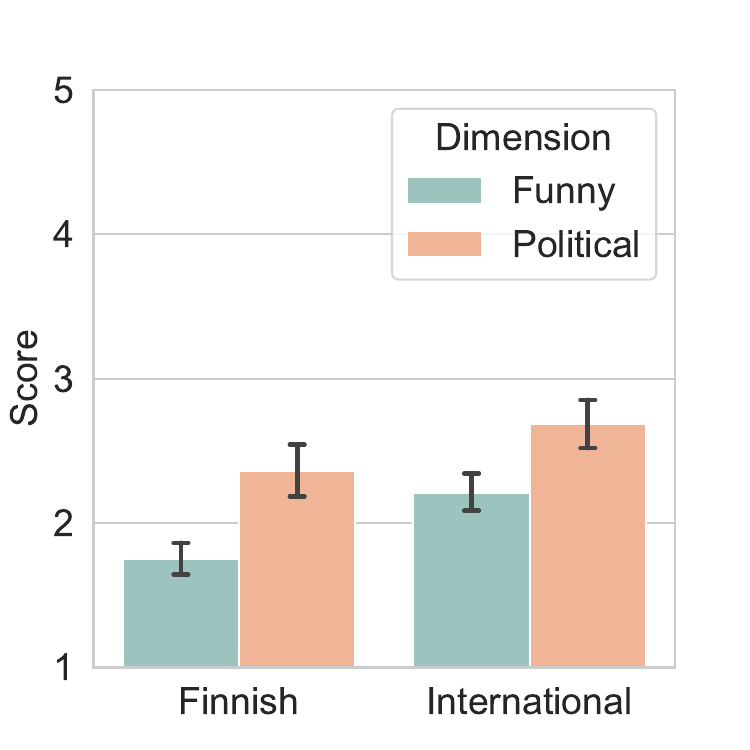}
        \caption{Finnish vs. International annotators}
        \label{fig:culture}
    \end{subfigure}\hfill
    \begin{subfigure}[t]{0.33\textwidth}
        \centering
        \includegraphics[width=\linewidth]{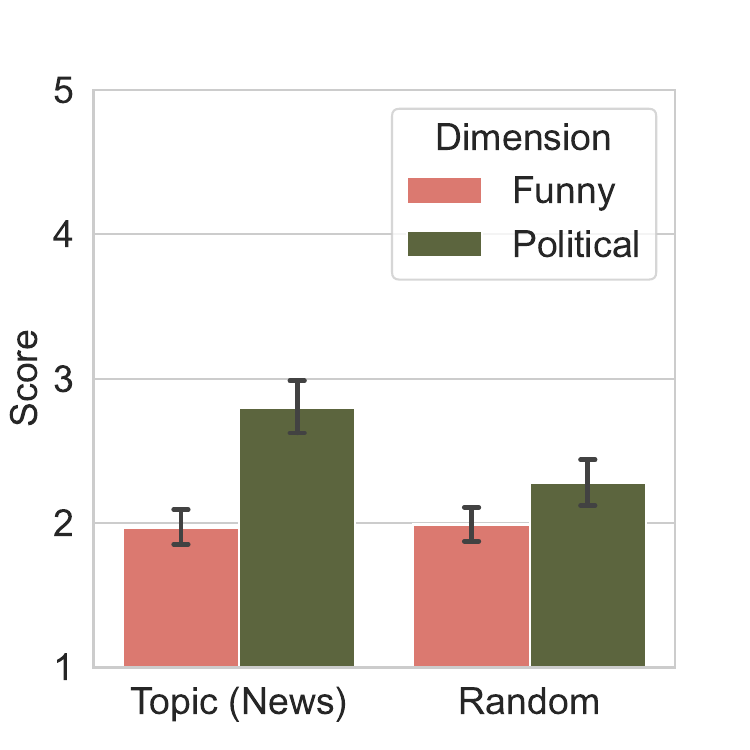}
        \caption{Topic vs Random words}
        \label{fig:topic-words}
    \end{subfigure}\hfill
    \begin{subfigure}[t]{0.33\textwidth}
        \centering
        \includegraphics[width=\linewidth]{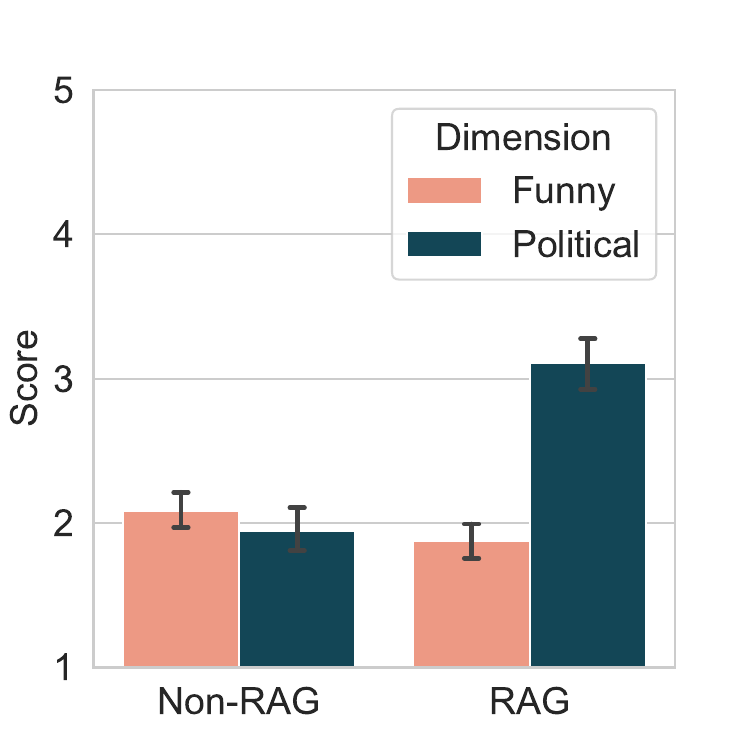}
        \caption{Non-RAG vs. RAG-based definitions}
        \label{fig:rag_dist}
    \end{subfigure}
    \caption{Average scores across experimental conditions.}
    \label{fig:score_distributions}
\end{figure*}

\paragraph{RQ1: To what extent are the generated definitions humorous and politically meaningful?}
Figure \ref{fig:score_distributions} shows the percentage distribution of scores for the two proposed questions. The generated satirical definitions are not perceived as funny by human annotators (M=1.98, SD=1.06), with 40\% of the annotations receiving a score of 1. They receive slightly higher ratings for political relevance (M=2.53,  SD=1.55) and a more diverse distribution of scores, which indicates that the definitions are perceived as more political than funny. For both dimensions, we observe a large standard deviation, which reflects substantial variability in the judgments. Variation in the ratings for both questions is expected, as both humor and political quality are subject to individual interpretation. This confirms that satire annotation is a hard task for humans.

\paragraph{RQ2: To what extent is successful satire generation dependent on cultural context? }

Table \ref{tab:IAA} reports inter-annotator agreement in the human evaluation, measured using Krippendorff’s $\alpha$. For the funny dimension, agreement is very low, which is consistent with the observations above and supports the view that humor perception is highly subjective. For the political dimension, agreement is somewhat higher, with an overall agreement above 0.5; however, this level should still be interpreted cautiously, as it falls below commonly used thresholds for strong reliability.

When considering the annotator groups separately, the international group shows slightly higher agreement on humor, whereas the Finnish group shows slightly higher agreement on political relevance. One possible explanation is that Finnish annotators may have been more familiar with current news topics, political discourse, and locally grounded satire, which could have led to more consistent judgments regarding what counts as political and contextually relevant in the Finnish setting. 

Figure \ref{fig:culture} shows the mean scores for both annotation questions. Contrary to our expectations, comparisons between Finnish and international annotators revealed no statistically significant differences in either of the ratings (p>0.1, Mann-Whitney U test). These results indicate that, in this dataset, cultural background did not have a systematic effect on the ratings.

\begin{table*}[t]
\adjustbox{max width=\textwidth}{
    \centering
    \begin{tabular}{lrrrr}
    \toprule
    & \multicolumn{2}{c}{Average Score} & \multicolumn{2}{c}{Human Correlations} \\
Model&	\multicolumn{1}{c}{Funny} & \multicolumn{1}{c}{Political}  & \multicolumn{1}{c}{Funny} & \multicolumn{1}{c}{Political} \\
\midrule
Aya-Expanse-8B	& 3.83 ± 0.77 & 3.40 ± 1.32 & \textbf{0.199* [0.005, 0.373]} & \textbf{0.826** [0.758, 0.872]} \\
EuroLLM-9B-Instruct	&  3.46 ± 0.81 & 2.41 ± 1.74 &  0.161  [-0.035, 0.334] &  0.663** [0.534, 0.760] \\
Llama-3.1-8B-Instruct	& 3.96 ± 0.57 & 3.21 ± 1.89 & 0.084 [-0.119, 0.265] & 0.756**  [0.671, 0.825]  \\
Mistral-7B-Instruct-v0.3 &  2.81 ± 0.67	& 1.83 ± 1.06 & -0.065 [-0.261, 0.134] & 0.751** [0.669, 0.816] \\
Qwen2.5-7B-Instruct	 & 3.52 ± 0.64 & 3.50 ± 1.50 & 0.069 [-0.145, 0.263] & 0.688** [0.580, 0.772] \\
\bottomrule
    \end{tabular}}
    \caption{Mean scores (\(M \pm SD\)) assigned by the LLM judges for the funny and political dimensions, together with Spearman correlation $\rho$ with human mean scores. Brackets indicate 95\% confidence intervals.  ** \(p < 0.001\), * \(p < 0.05\).}
    \label{tab:llms}
\end{table*}

\paragraph{RQ3:  How does the choice of the candidate word affect the quality of the generated definitions?}
We conduct a Mann-Whitney U test to compare annotations for randomly selected words and topic-modeled candidate words. The results show no statistically significant difference for the funny dimension (p=0.758), whereas the political dimension exhibits a statistically significant difference (p<0.001). These findings indicate that words selected through topic modeling lead to definitions that are perceived as more political, but not funnier, than definitions based on randomly selected words.

\paragraph{RQ4: Does RAG improve the quality of generated satirical definitions?}
To compare annotations for definitions generated with and without RAG, we conduct a Wilcoxon signed-rank test. Figure \ref{fig:rag_dist} presents the mean scores for this comparison. As in the previous analysis, RAG does not yield a statistically significant difference in the funny dimension (p=.05), but it does lead to a statistically significant improvement in the political dimension (p<.001). These results indicate that our RAG pipeline is more effective than the non-RAG baseline at generating politically relevant content, but not at improving humor. This outcome is in line with our expectations, since the purpose of grounded generation is to anchor outputs in the provided source material, and in our case the retrieved news snippets are not themselves expected to be humorous or satirical.

\begin{figure}[]
    \centering
    \begin{subfigure}[t]{\linewidth}
        \centering
        \includegraphics[width=\linewidth]{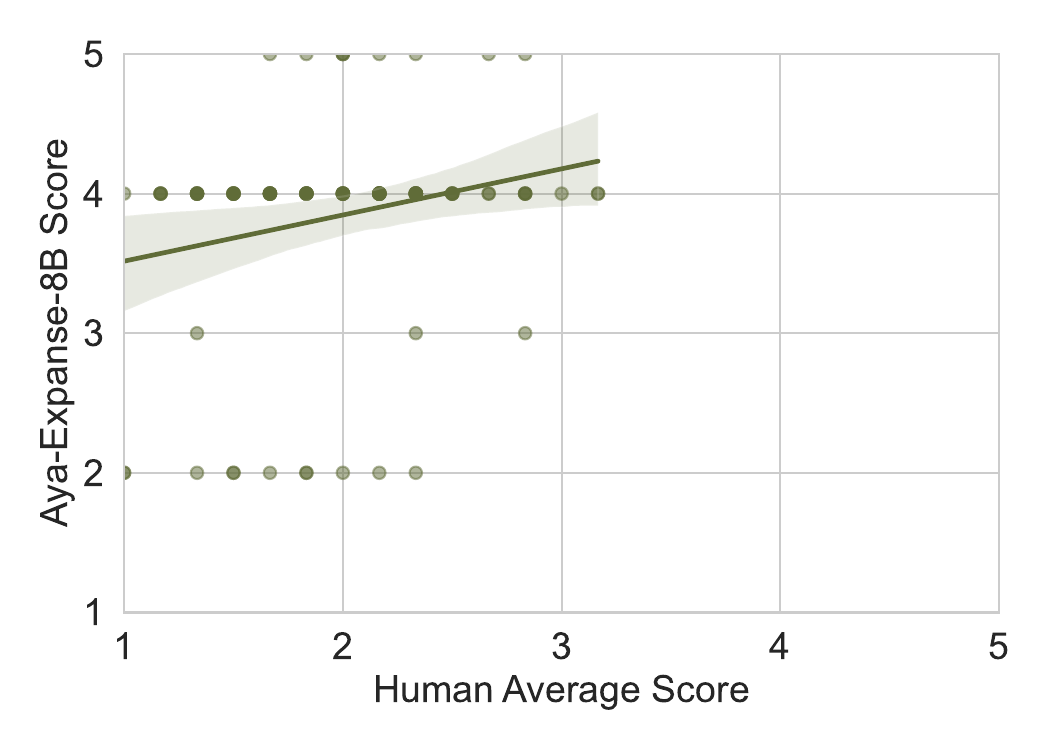}
        \caption{Funny}
        \label{fig:aya-funny}
    \end{subfigure}\hfill
    \begin{subfigure}[t]{\linewidth}
        \centering
        \includegraphics[width=\linewidth]{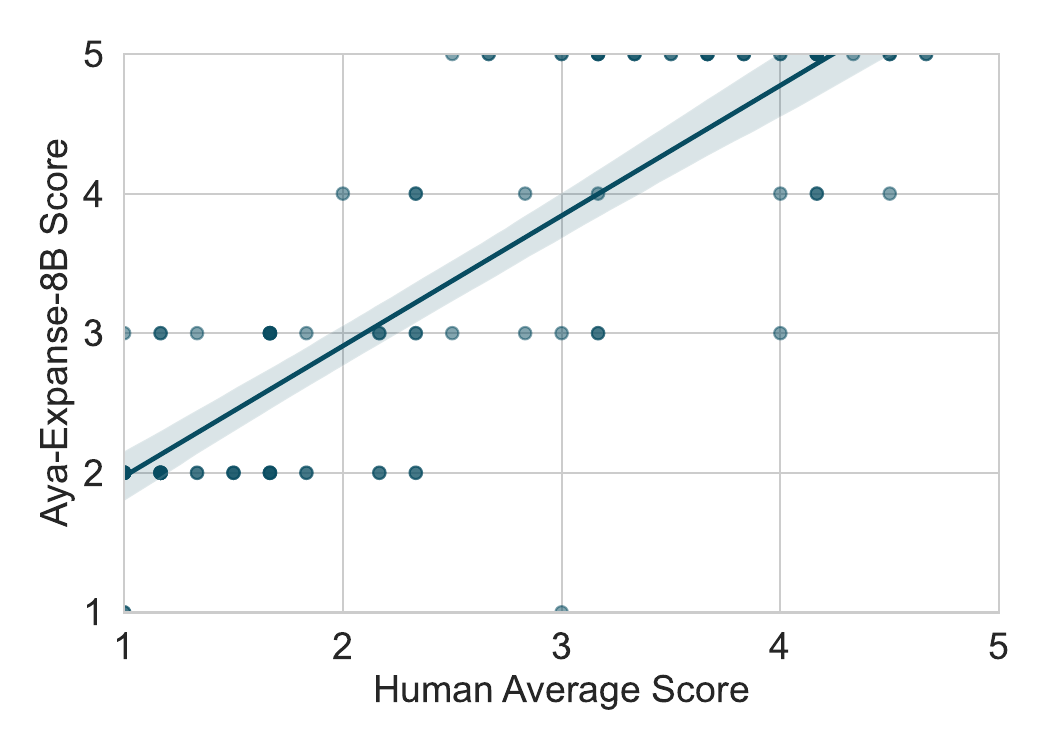}
        \caption{Political}
        \label{fig:aya-political}
    \end{subfigure}
    \caption{Correlation of human scores with Aya-Expanse-8B annotations.}
\end{figure}

\paragraph{RQ5: Can LLMs serve as reliable evaluators of satirical content?}
Table \ref{tab:llms} reports the mean scores assigned by each LLM judge, together with their correlations with human ratings. Figures \ref{fig:aya-funny} and \ref{fig:aya-political} present the score distributions for Aya-Expanse-8B and its correlation with human judgments, while Appendix \ref{sec:corr-llms} provides the corresponding figures for the remaining four models. 

Overall, the LLMs assign higher mean scores to humor, with relatively low variance, and lower mean scores to political relevance, with greater variability. 
Based on the correlation scores, the evaluated LLMs do not capture humor well, as their correlations with human ratings on the funny dimension are uniformly low. By contrast, all models show strong correlations with human judgments on the political dimension, indicating that they are much better at identifying political relevance than humor. Among the evaluated models, Aya-Expanse-8B achieves the highest correlation with human judgments overall.

Taken together, these results indicate that LLMs can serve as reasonably reliable evaluators of the political relevance of satirical definitions, but they remain poor judges of subjective qualities such as humor.

\section{Web Application}
\label{UI}

To showcase our pipeline we built a web application. It can be run locally to generate definitions and search for relevant news with any user input, e.g. keywords presented in a plotting based on topic modeling. The repository of the application will be made public upon acceptance. 

\section{Conclusions}

In this study, we presented a novel pipeline that uses RAG for grounded satire generation. We formulated five research questions and evaluated the system through both human annotation and an LLM-as-a-judge framework.

Our results show that the generated definitions are perceived as more political than humorous. Contrary to our expectations, considering satire as a culturally dependent phenomenon, we do not find a statistically significant effect of cultural background on annotation outcomes in our dataset. 
We found that both RAG and topic-based word selection improve the political relevance of the generated content, indicating that our pipeline is effective for generating politically grounded satire. However, neither leads to clear improvements in humor generation. Finally, the results of our LLM-as-a-judge evaluation show that LLMs are effective at identifying politically relevant content, but perform poorly at detecting humor. This further supports the view that humor remains a particularly difficult and subjective evaluation task for current language models.

Our current pipeline is limited to English news from the Finnish public broadcaster Yle. Future work should extend the approach to additional languages and sources, and explore alternative evaluation methods in order to provide a broader and more robust assessment of system performance.

\bibliography{custom}

\appendix

\section{Prompts for Generating Definitions}
\label{sec:prompts}

\subsection{Prompt with RAG}

\begin{lstlisting}
You are the editor of a 'Satirical Dictionary'. Define the term based SOLELY on the provided news context.
CRITICAL RULE: You must use the SPECIFIC IRONY found in the text, not generic stereotypes.
Example: If the text says 'working people need food', do NOT joke about laziness. Joke about how wages are useless.
Style Guidelines:
1. Cynical and Dark.
2. Highlight the absurdity of the specific situation described in the text.
3. ATTENTION: Keep it under 50 words.
Only output the definition, No explanations or commentary.
\end{lstlisting}

\subsection{Prompt without RAG}

\begin{lstlisting}
You are the editor of a 'Satirical Dictionary'.
CRITICAL RULE: You must use SPECIFIC IRONY typical for Finnish culture.
Style Guidelines:
1. Cynical and Dark.
2. ATTENTION: Keep it under 50 words.
Only output the definition, No explanations or commentary.
\end{lstlisting}

\subsection{Prompt for LLM-as-a-judge}

\begin{lstlisting}
Task:
Score a satirical definition on two dimensions:
- funny
- political

Use only the text provided by the user.
Do not use external knowledge.
Do not explain your answer.
Do not add any text before or after the JSON.

Scales:

funny:
1 = not funny
2 = slightly funny
3 = funny
4 = very funny
5 = extremely funny

political:
1 = not political
2 = slightly political
3 = generally political
4 = clearly political and topical
5 = strongly political and specifically relevant to Finnish political culture

Output rules:
- Output exactly one JSON object
- Use exactly these two keys: "funny", "political"
- Both values must be integers from 1 to 5
- Do not use markdown
- Do not use code fences
- Do not output anything except the JSON object

Valid output example:
{"funny": 3, "political": 4}
"""
\end{lstlisting}

\begin{comment}
    \section{Annotation guidelines}
”There are 100 words and their definitions. Your task is to evaluate the definition based on two questions: Q1: Is it funny? and Q2: Is it political?. The evaluation scale is from 1-5 and below you can find verbal explanations of each score for both questions, respectively.

\begin{table}[h]
  \centering
  \begin{tabular}{ll}
    \hline
    \textbf{Q1: Is it funny?} \\
    \hline
    \textbf{Grade} & \textbf{Explanation} \\
    1   & not funny, awkward at most \\
    2   & slightly funny\\
    3   & so funny I laughed \\
    4   & funny, and I will tell it to  \\
       & someone else \\
    5  & so funny I will laugh if       \\
       & I tell it to someone else    \\
    \hline
  \end{tabular}
  \caption{Annotation guideline for Q1: \textit{Is it funny?}}
  \label{tab:Q1}
\end{table}

\begin{table}[h]
  \centering
  \begin{tabular}{ll}
    \hline
    \textbf{Q2: Is it political?} \\
    \hline
     \textbf{Grade} & \textbf{Explanation} \\
    1   & not political at all \\
    2   & some political quality\\
    3   & political on a general level \\
    4   & political and current  \\
    5  & political, current and rele-  \\
       & vant in Finnish political   \\
       & culture \\
    \hline
  \end{tabular}
  \caption{Annotation guideline for Q2: \textit{Is it political?}}
  \label{tab:Q2}
\end{table}

\end{comment}

\section{Correlations with Human Judgements}

\begin{figure*}[h]
    \centering
    \begin{subfigure}[t]{0.4\textwidth}
        \centering
        \includegraphics[width=\linewidth]{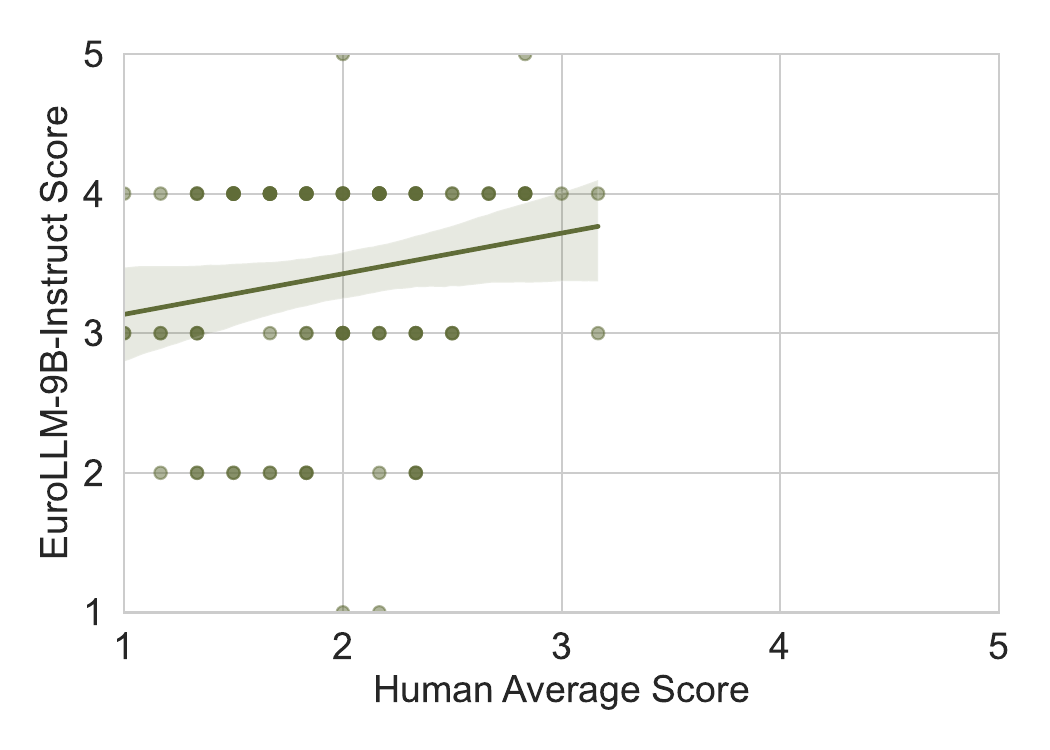}
        \caption{Funny}
    \end{subfigure}\hfill
    \begin{subfigure}[t]{0.4\textwidth}
        \centering
        \includegraphics[width=\linewidth]{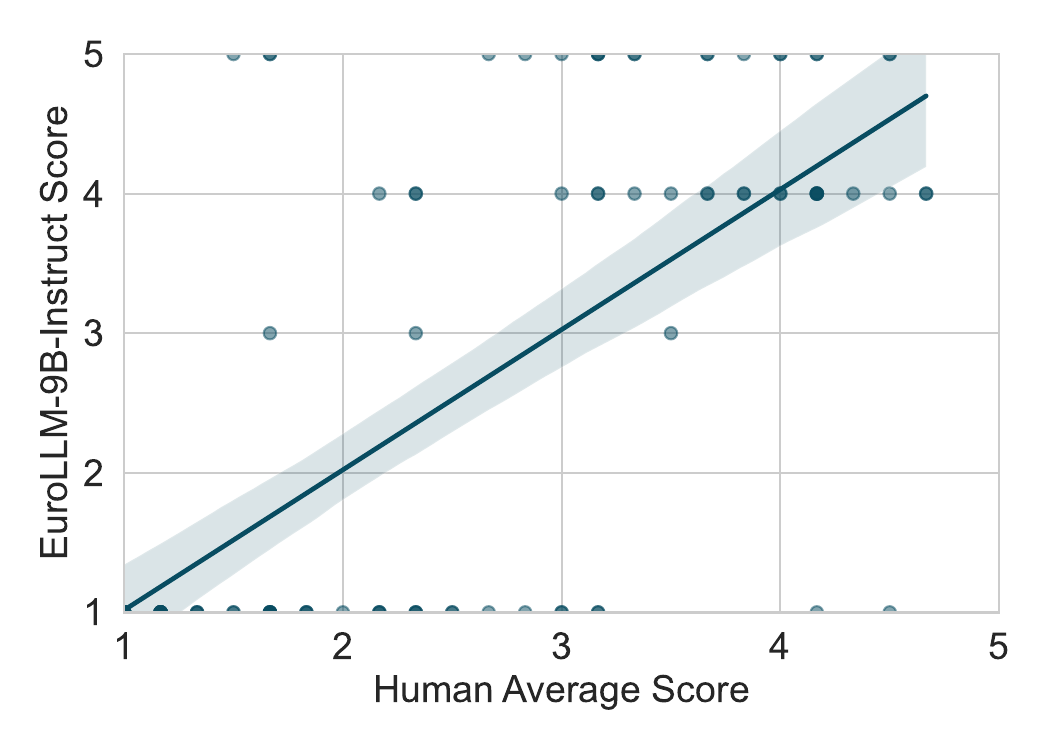}
        \caption{Political}
    \end{subfigure}\hfill
    \caption{Correlation of human scores with EuroLLM-9B-Instruct Annotations}
    \label{fig:eurollm}
\end{figure*}

\begin{figure*}[h]
    \centering
    \begin{subfigure}[t]{0.4\textwidth}
        \centering
        \includegraphics[width=\linewidth]{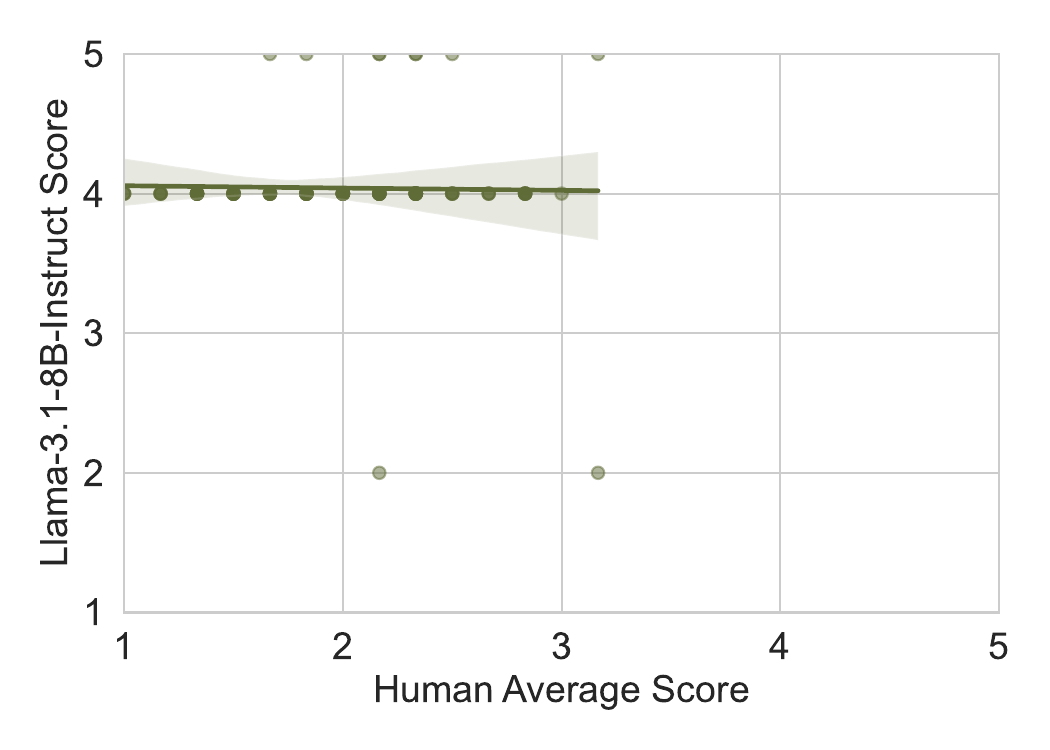}
        \caption{Funny}
    \end{subfigure}\hfill
    \begin{subfigure}[t]{0.4\textwidth}
        \centering
        \includegraphics[width=\linewidth]{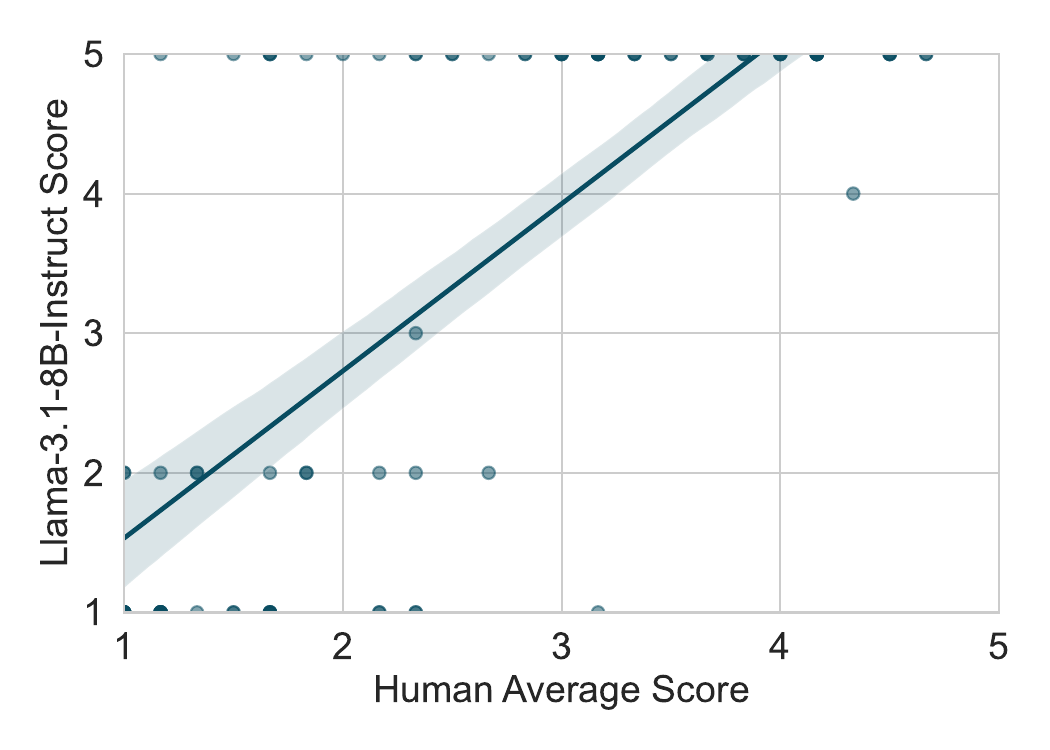}
        \caption{Political}
    \end{subfigure}\hfill
    \caption{Correlation of human scores with Llama-3.1-8B-Instruct Annotations}
    \label{fig:llama}
\end{figure*}

\begin{figure*}[h]
    \centering
    \begin{subfigure}[t]{0.4\textwidth}
        \centering
        \includegraphics[width=\linewidth]{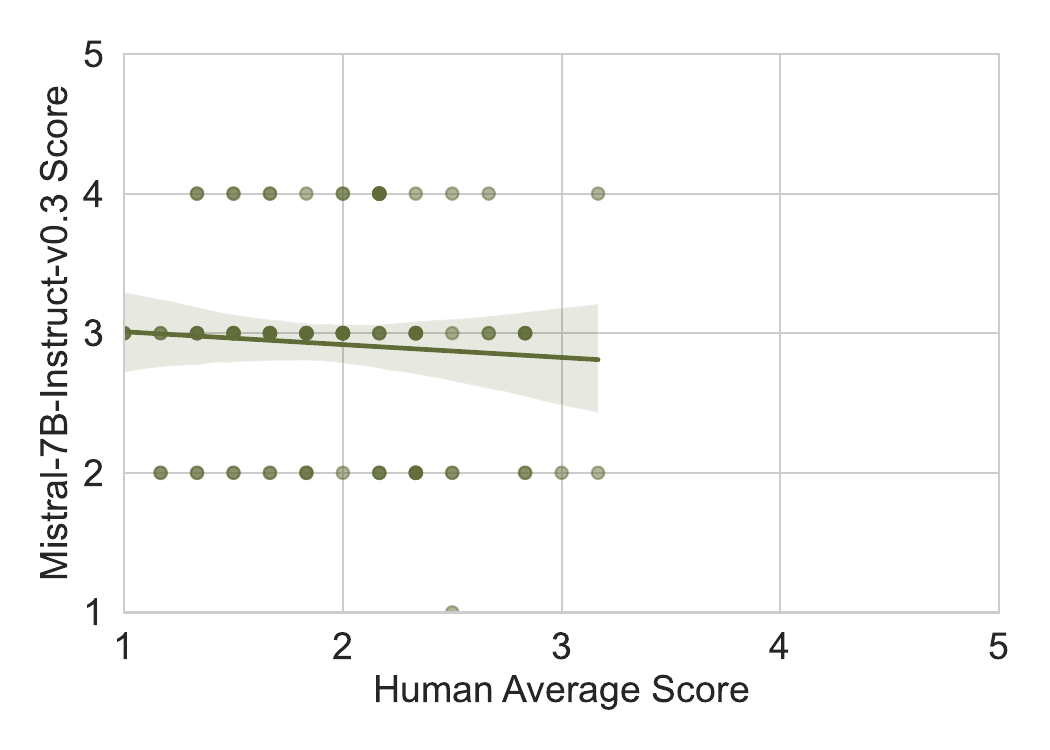}
        \caption{Funny}
    \end{subfigure}\hfill
    \begin{subfigure}[t]{0.4\textwidth}
        \centering
        \includegraphics[width=\linewidth]{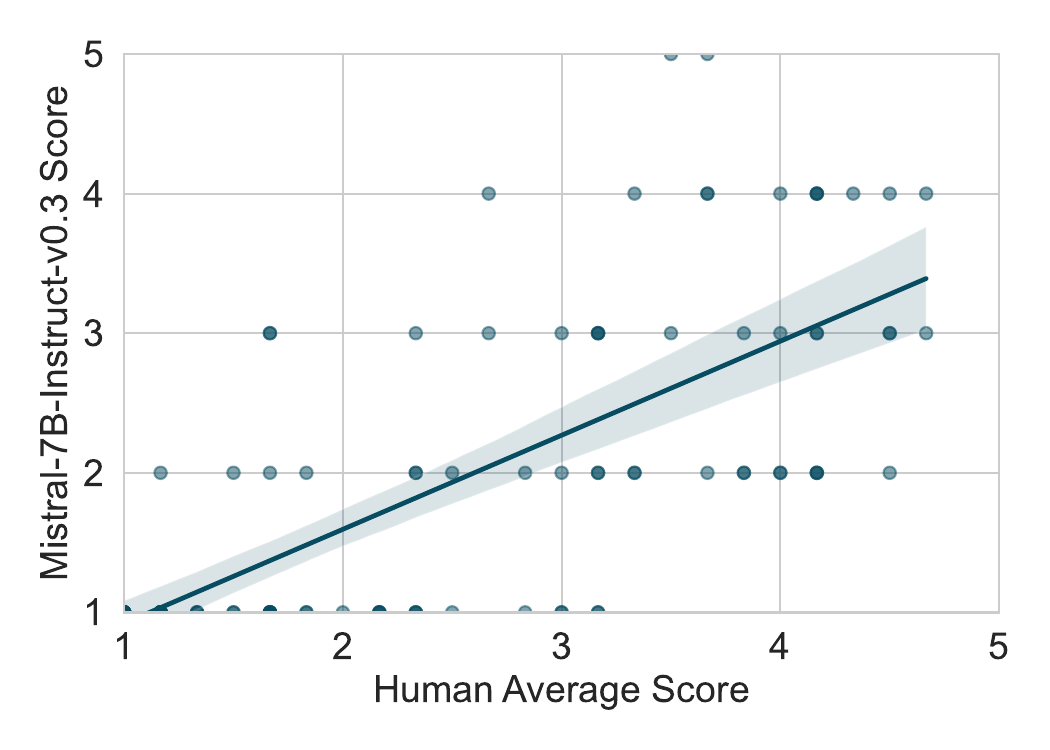}
        \caption{Political}
    \end{subfigure}\hfill
    \caption{Correlation of human scores with Mistral-7B-Instruct Annotations}
    \label{fig:mistral}
\end{figure*}

\begin{figure*}[h]
    \centering
    \begin{subfigure}[t]{0.4\textwidth}
        \centering
        \includegraphics[width=\linewidth]{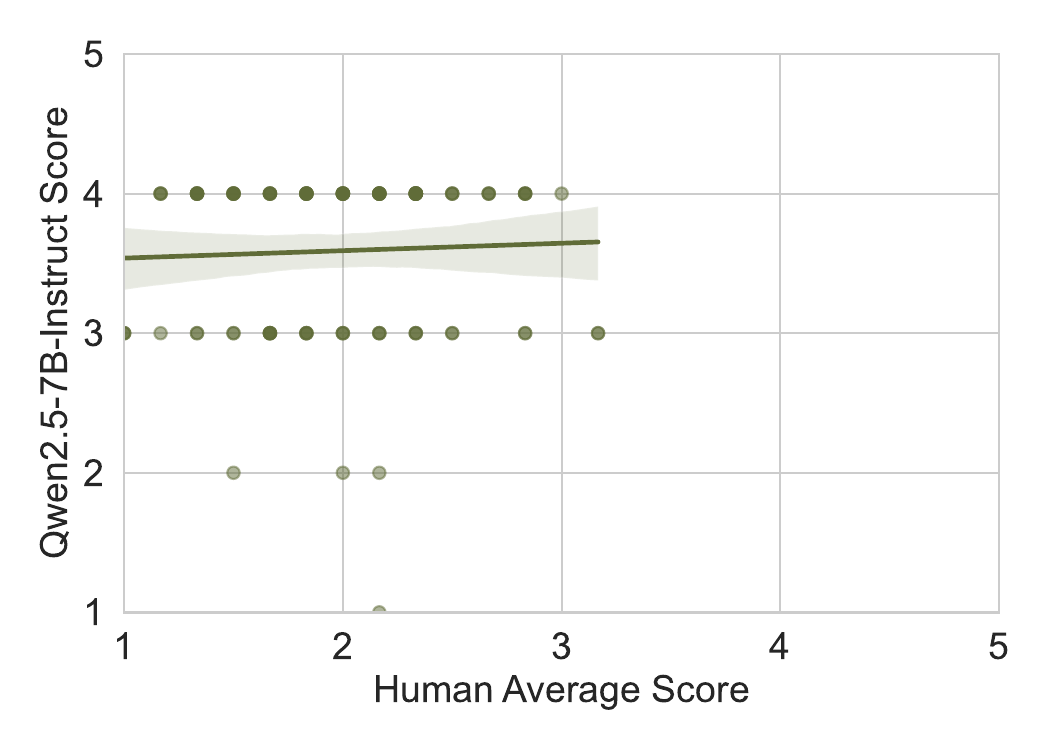}
        \caption{Funny}
    \end{subfigure}\hfill
    \begin{subfigure}[t]{0.4\textwidth}
        \centering
        \includegraphics[width=\linewidth]{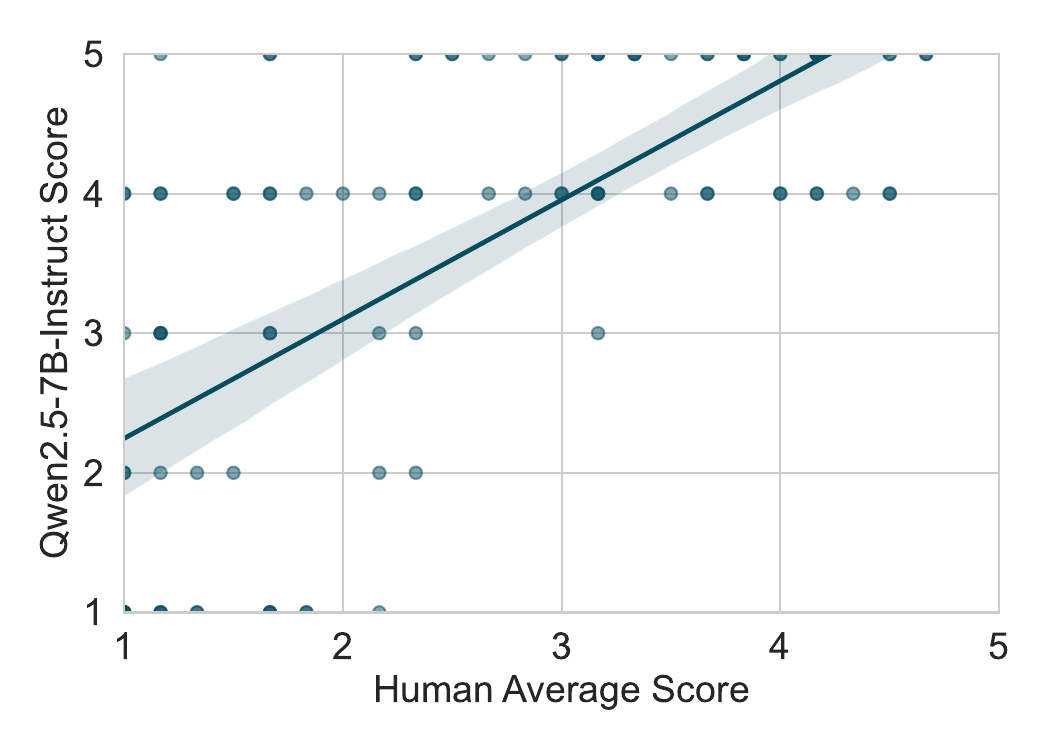}
        \caption{Political}
    \end{subfigure}\hfill
    \caption{Correlation of human scores with Qwen2.5-7B-Instruct Annotations}
    \label{fig:qwen}
\end{figure*}

Figures \ref{fig:eurollm}, \ref{fig:llama}, \ref{fig:mistral} and \ref{fig:qwen} show the correlations of LLM judgments with human annotations.

\label{sec:corr-llms}

\end{document}